\documentclass[conference]{IEEEtran}
\IEEEoverridecommandlockouts
\usepackage{cite}
\usepackage{amsmath,amssymb,amsfonts}
\usepackage{algorithmic}
\usepackage{graphicx}
\usepackage{textcomp}
\usepackage{xcolor}
\usepackage{comment}

\def\BibTeX{{\rm B\kern-.05em{\sc i\kern-.025em b}\kern-.08em T\kern-.1667em\lower.7ex\hbox{E}\kern-.125emX}}

\begin{document}

%Bangla Sarcasm Detection with BERT using Stratified K-Fold Cross Validation
\title{Interpretable Bangla Sarcasm Detection using BERT and Explainable AI\\
}

\author{\IEEEauthorblockN{Ramisa Anan}
\IEEEauthorblockA{\textit{Computer Science and Engineering} \\
\textit{BRAC University}\\
Dhaka, Bangladesh \\
ramisa.anan@g.bracu.ac.bd}
\and

\IEEEauthorblockN{Tasnim Sakib Apon}
\IEEEauthorblockA{\textit{Computer Science and Engineering} \\
\textit{BRAC University}\\
Dhaka, Bangladesh \\
sakibapon7@gmail.com}
\and

\IEEEauthorblockN{Zeba Tahsin Hossain}
\IEEEauthorblockA{\textit{Computer Science and Engineering} \\
\textit{BRAC University}\\
Dhaka, Bangladesh \\
zeba2tahsin@gmail.com}
\and

\IEEEauthorblockN{ Elizabeth Antora Modhu}
\IEEEauthorblockA{\textit{Computer Science and Engineering} \\
\textit{BRAC University}\\
Dhaka, Bangladesh \\
elizabeth.antora.modhu@g.bracu.ac.bd}
\and

\IEEEauthorblockN{Sudipta Mondal}
\IEEEauthorblockA{\textit{Computer Science and Engineering} \\
\textit{BRAC University}\\
Dhaka, Bangladesh \\
mondal.sudiptahere@gmail.com}
\and

\IEEEauthorblockN{MD. Golam Rabiul Alam}
\IEEEauthorblockA{\textit{Computer Science and Engineering} \\
\textit{BRAC University}\\
Dhaka, Bangladesh \\
rabiul.alam@bracu.ac.bd}
}

\maketitle

\begin{abstract}
    A positive phrase or a sentence with an underlying negative motive is usually defined as sarcasm that is widely used in today's social media platforms such as Facebook, Twitter, Reddit, etc. In recent times active users in social media platforms are increasing dramatically which raises the need for an automated NLP-based system that can be utilized in various tasks such as determining market demand, sentiment analysis, threat detection, etc. However, since sarcasm usually implies the opposite meaning and its detection is frequently a challenging issue, data meaning extraction through an NLP-based model becomes more complicated. As a result, there has been a lot of study on sarcasm detection in English over the past several years, and there's been a noticeable improvement and yet sarcasm detection in the Bangla language's state remains the same. In this article, we present a BERT-based system  that can achieve 99.60\% while the utilized traditional machine learning algorithms are only capable of achieving 89.93\%. Additionally, we have employed Local Interpretable Model-Agnostic Explanations that introduce explainability to our system. Moreover, we have utilized a newly collected bangla sarcasm dataset, BanglaSarc that was constructed specifically for the evaluation of this study. This dataset consists of fresh records of sarcastic and non-sarcastic comments, the majority of which are acquired from Facebook and YouTube comment sections. 
    %However, sarcasm detection in dialogues has become a challenge in the Natural Language Processing (NLP) field. In this research, we are building a system which can detect sarcasm effectively using Various machine learning model. For the classification task, it utilizes the Machine Learning techniques and RNN (Recurrent Neural Networks) for the extraction features. The main goal is to reduce the amount of time it takes humans to extract sarcasm from reviews and to develop a model that can recognize sarcasm in natural language. 
\end{abstract}

\begin{IEEEkeywords}
Machine Learning, Natural Language Processing, Sarcasm Detection, BERT.

\end{IEEEkeywords}

\section{Introduction}

    %Sarcasm is the use of words that signify the exact opposite of whatever we are trying to imply, usually to insult or mock someone, demonstrate irritation, or just be humorous. The advent of the internet changed the way people connect all around the world, giving the world a new perspective. Now-a-days, sarcasm is  deep-rooted throughout the internet all over the world. We use sarcasm in face to face conversation as well as in the chat box or comment section of social media platforms.
    %Sentiment analysis research has enabled machines to identify whether a text is positive, negative, or neutral with a high degree of accuracy, depending on the dataset. Finding the exact sentiment is incredibly difficult when the current phrase is coated with sarcasm, making it extremely impossible to determine if the line is expressed in a sarcastic manner or not. Hence, the significance of sarcasm detection for sentiment analysis, as well as the problems in identifying it, turn out to be a research problem. on the other hand, is a challenging endeavor as it relies heavily on context along with past information, and the tone with which the line was stated or written.

    Sarcasm is the use of words that mean the complete opposite of what we are attempting to convey and most of the time it is used in order to mock, criticize, or express discontent. Sarcasm is ingrained in today's online culture all around the world and is used often in verbal exchanges in person as well as in chat rooms and comment sections of social networking sites. Automated sarcasm detection algorithms can be beneficial for various NLP applications, including marketing analysis, opinion mining, and information classification. \par
    Sarcasm often alters the polarity of a remark, making its detection and processing crucial in an automated NLP system. In recent times, a  great deal of effort has been put towards automating the sarcasm detection process since it allows for more accurate analytics in online comments and reviews. Sentiment and emotion analysis approaches are being used to study predicting and removing phrases that suggest sarcasm with the growth of artificial intelligence (AI) tools, notably NLP and as a result sarcasm detection in the English language has improved a lot in the past few years. However, for the Bangla language, it still remains a challenging task, and a very limited effort has been made to address this fundamental problem. The primary objectives of our investigation are as follows: (i) Introduce an Automated sarcasm detection system in the Bangla language that can aid in various NLP-based systems such as sentiment analysis, opinion mining, and marketing tools. (ii) Increase the efficiency of the entire process by performing a thorough study in order to find out the optimal framework. While conducting we encountered some challenges. The availability of Bangla sarcasm data was the most notable along with the limitations of previous studies. \par
    
    The contributions of this article are as follows:
    \begin{itemize}
        \item A Bidirectional Encoder Representations from Transformers(BERT) based model with Stratified K-fold Cross Validation has been proposed for detecting sarcasm. A comprehensive study of various traditional machine learning models with the proposed system is presented as well.
        \item Local Interpretable Model-Agnostic Explanations is utilized in order to explore the interpretability of the proposed model.

        \item To assess the effectiveness of this study's proposed model, we have utilized the BanglaSarc dataset, a fresh dataset on sarcasm. This dataset is collected from Facebook, YouTube, and various online resources for the validation of the proposed model specifically. 
        
        %\item 
    \end{itemize}
    In our article, section II briefly covers recent work in the field of sarcasm detection. The dataset has been used to train our model as described in section III. Section IV demonstrates the process of the model. The results and analysis have been illustrated in Section V. Finally, section VI concludes with a hope for improved performance of our model and future work in the same sector.

\section{RELATED WORK}
    Sarcasm analysis has caught the curiosity of the academic community for a long time. Many studies on sarcasm detection have lately been published, with the majority of them focusing on English and other Indo-European languages \cite{Lunando}. The majority of sarcasm detection research has been done on textual data by evaluating lexical features. As NLP is a field that primarily focuses on evaluating textual documents, it is vastly useful in terms of sarcasm detection. \par
    For instance, Razali et al. used a Convolution Neural Network (CNN) to extract deep features from sarcastic tweets, with an Artificial Neural Network (ANN) as their training approach \cite{Razali}. Their research centered on the detection of sarcasm in tweets by combining deep learning-derived features with contextual handcrafted information. 780,000 English tweets (130,000 sarcastic and 650,000 non-sarcastic) were distributed unevenly across their produced dataset. The dataset used for their research was divided into training and testing portions, respectively, of 80\% and 20\%. With all of the feature sets combined, they performed an experiment using Logistic Regression as a classification technique along with other techniques. They discovered that it was the most effective method for the task. However, it does not perform as well as a stand-alone function. \par
    A context-based feature strategy based on a deep learning model and standard machine learning was proposed by Eke et al.\cite{Eke}. To make the model training process more efficient, they combined the Internet Argument Corpus, version two (IAC-v2) benchmark dataset with two Twitter datasets: the Riloff dataset (1956 tweets) and the Ghosh and Veale dataset (54929 tweets). For learning context and word embedding, the author combined Bi-LSTM, a type of RNN, and the BERT model with Global Vector Representation (GloVe). When compared to the current approaches, their third model's(Proposed Feature Fusion model) average detection precision ranged from 3.7 to 10.2 percent and was based on feature fusion using the BERT feature, sentiment-related, syntactic, and GloVe embedding features. \par
    Jamil et al. combined CNN and LSTM networks to recognize sarcasm in input by looking at it as a sequence of embedded words \cite{Jamil}. Its foundation is based on a deep neural network (DNN), which can mimic biological neurons and conduct complex computational modeling. They have used multiple datasets in this study including the Tweet dataset(39,267 tweets) and News Headlines dataset(26,709) for training and testing purposes also the Reddit dataset(15,000) and Sarcasm Corpus V2(6,520) for validation. With an accuracy of 91.60 \%, the experimental findings showed that the suggested model outperformed the capabilities of the standard machine learning algorithms. \par
     Kumar et al. proposed one of the most widely used algorithms for detecting sarcasm in conversations, with the primary goal of anticipating sarcasm in each speaker's statement \cite{Kumar}. They used the MUStARD dataset, which contains 690 lines of conversation from four well-known sitcoms. There are an equal amount of sarcastic and non-sarcastic exchanges among the data. They used the Python framework sci-kit-learn to create the XGBoost learning model. Accuracy, F1 score, precision, and recall were used as metrics to assess XGBoost's classification performance.  \par
    Another ensemble model developed by the authors Das et al. proposed a sarcasm detection model with four Parallel Long Short Term Memory (pLSTM) networks and softmax activation function and got the best overall validation accuracy of 98.31 percent \cite{Das}. They chose the MUStARD or Multimodal Sarcasm Detection Dataset's final textual corpus, which consists of 690 lines of individually indexed dialogues, to train their model on. The dataset included the initial token average for the sentences as well as the instances of sarcastic word utterances that are part of the relevant sentences.  \par
    Additionally, LIME has been employed so that users may understand how predictions are created \cite{lime}. LIME is broadly used in interpreting various deep learning-based frameworks that are used in both image and text-based tools \cite{Zucco} \cite{apon} \cite{Ye}. \par
    Unlike the studies mentioned above, we generated a Bangla Dataset specifically for this research and utilized a traditional machine-learning model along with BERT in order to classify sarcasm.

    \begin{table}[!t]
        \caption{Comparison of various studies on sarcasm detection}
            \begin{center}
                    \begin{tabular}{|c |c |c |c |}
                        \hline
                         Model & Dataset & Accuracy & Ref  \\
                        \hline
                        Logistic Regression     &  English Tweets &  94\%   &\cite{Razali}\\ 
                        \hline
                        Proposed Feature Fusion &  English Tweets &  76.3\% &\cite{Eke}\\
                        \hline
                         CNN + LSTM & News Headlines & 73\%  &\cite{Jamil}\\
                         \hline
                         XGBoost  & MUStARD  & 93.1\%  &\cite{Kumar}\\
                         \hline
                         LSTM   & MUStARD & 98.5\% &\cite{Das}\\
                         \hline
                    \end{tabular}
                    \label{fig:x RelatedWorkTable}
            \end{center}
    \end{table}

\section{METHODOLOGY}
    In this section, we get an overview of our proposed sarcasm detection model, which is separated into three sub-sections. Sub-section \ref{proposedModel} of this paper discusses our proposed model. However, Sub-section \ref{Data} is divided into two sub-subsections. In the initial sub-subsection we discuss about our data acquisition techniques and description and in the later sub-subsection we discuss data pre-processing. Finally,  \ref{modelSpecification} exposes our model's framework.
    
    \begin{figure*}[!t]
        \centering
        \includegraphics[scale= 1.02]{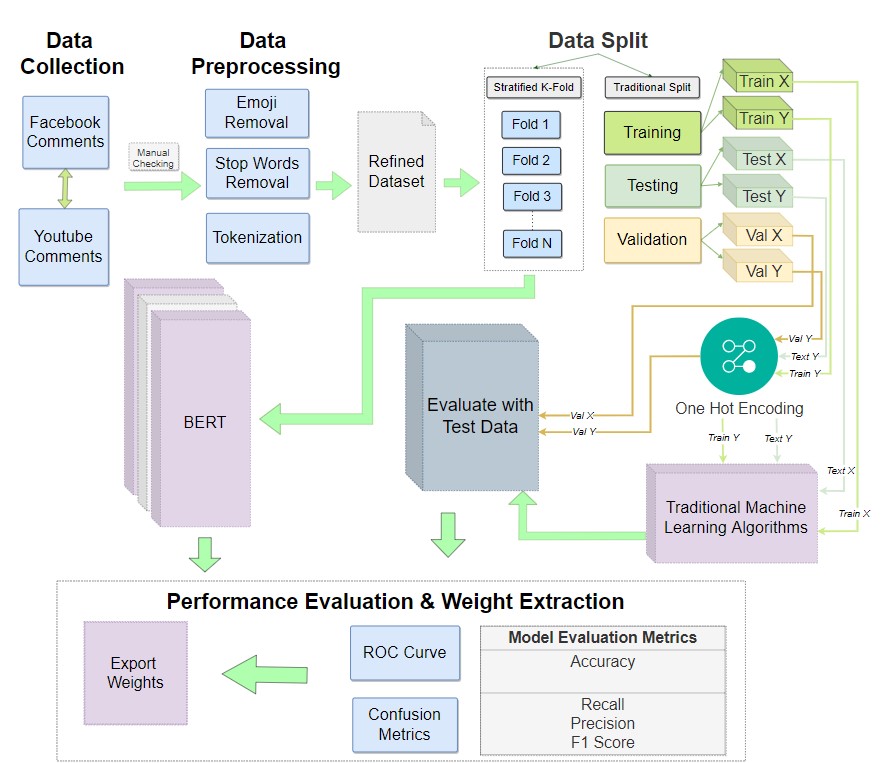}
        \caption{System Model: Bangla Sarcasm Detection System Model outlining the overall process.}
        \label{fig:x proposedmodels}
    \end{figure*}

\subsection{Proposed Model} \label{proposedModel}

     Initially, we started by gathering data from various social media platforms namely Facebook, YouTube, etc. Since collected data from social platforms tend to be very noisy, we had to consider a series of pre-processing techniques in order to clean the data. However, before doing so, we manually examined and labeled each of the data to confirm that there are no discrepancies between the data. After performing manual checking and pre-processing techniques over our gathered dataset, it proved to be advantageous. Finally, before feeding our data to our preferred model, we have split the data into two parts which are stratified K-Fold and traditional split where stratified K-Fold is used with BERT. Furthermore, the traditional split is divided into three sections which are train, test, and validation whereas the train set contains 60\% of the data and the other two sets have 20\% each. In addition, we have also performed one hot encoding on the labels for all three of the split sets. Later, we fed the train and test set into various traditional machine learning models namely, Logistic Regression, Decision Tree, Random Forest, Naive Bayes, K-Nearest Neighbor(KNN), Support Vector Machine(SVM), SGD, TFIDF along with K-Fold as Cross-Validation with a BERT. After training our model, with the validation set, we evaluate our trained model's performance. We have as well employed various performance metrics along with ROC Curve and confusion metrics. Here, we have performed a comprehensive comparison between the trained models from BERT and evaluated them with a test dataset in order to find out which model tends to be more efficient in terms of accuracy, memory, speed, etc finally we export the best-fitted model according to our analysis. With the exported model we have employed LIME (Local Interpretable Model-agnostic Explanations) that introduces explainability to our study. This explainability helps us understand how our model comprehends and classifies a prediction. 

%The total number of datasets is 5150 and the number ‘0’ denotes a non sarcastic statement, while ‘1’ denotes a sarcastic comment. After data pre-processing, our dataset left with 5112 data. Here, we found 3159 non-sarcastic opinions and 1953 sarcastic remarks. Based on this, we can conclude that the dataset is balanced.

% \cite{data_ref}
\subsection{Data: BanglaSarc :} \label{Data}
    \subsubsection{Data Acquisition \& Description}
       In this work, we utilized the BanglaSarc Dataset in order to evaluate our proposed model \cite{data_ref}. BanglaSarc Dataset is gathered with Bangla comments from Facebook, YouTube, and other online resources. Afterward, these data were manually reviewed and assigned labels for each of the collected records. The authors then excluded noisy data throughout the screening process. Authors further labeled the data with '0' and '1' where '0' implies a non-sarcastic statement and '1' denotes a sarcastic statement in this situation. However, these gathered datasets turned out to be rather noisy after all, which necessitated us to manually clean the data. Initially, BanglaSarc Dataset is consisted of 7800 comments however, after manually filtering authors had to discard almost 2700 records. As a result, the final dataset consisted of 5112 comments whereas it contains about 3159 non-sarcastic data and 1951 sarcastic data. Even though the BanglaSarc dataset is not entirely balanced, we may nevertheless draw the conclusion that its quality is moderate.

    \subsubsection{Data Pre-processing}
\begin{figure}[htbt!]
            \centering
            \includegraphics[scale= .6]{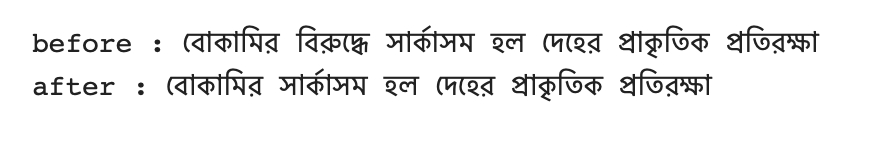}
            \caption{Data Pre-processing: Sample Input.}
            \label{Dataprocessing}
        \end{figure}

         We begin by removing all of the emojis as managing emojis and emoticons is essential when doing text pre-processing tasks in NLP. The stopwords are then removed since stopwords do not add much sense to a phrase. Generally, they are excluded from the dataset to improve model performance. We tokenized the text and then removed all the punctuation. Tokenization is the process of breaking down a piece of text into smaller chunks or units known as tokens. Data pre-processing is an important step in the development of a machine learning model. Figure \ref{Dataprocessing} represents an example input and output using our pre-processing methods.\par 

\subsection{Model Specification} \label{modelSpecification}

    \begin{table*}[htbt!]
        \caption{Performance Evaluation: Traditional Machine Learning Models}
            \begin{center}
                    \begin{tabular}{|c |c |c |c |c|}
                        \hline
                        Model & Accuracy & Precision & Recall & F-1 Score\\ [0.5ex]
                        \hline
                        Random Forest               & 89.93  & 89.93  & 89.93 & 89.93 \\
                        \hline
                        Decision Tree               & 83.58	 & 83.58  & 83.58 & 83.58 \\
                        \hline
                        K-Nearest Neighbor          & 74.78  & 74.78  & 74.78 & 74.78 \\
                        \hline
                        Support-Vector Machines     & 71.55 & 71.55 & 71.55 & 71.55 \\
                        \hline
                        Multinomial Naive Bayes     & 65.10 & 65.10 & 65.10 & 65.10 \\
                        \hline
                        Logistic Regression         & 62.46 & 62.46 & 62.46 & 62.46 \\
                        \hline
                        Stochastic Gradient Descent & 53.86 & 53.86 & 53.86 & 53.86 \\
                        \hline
                        BERT & 99.60  & 99.60  & 99.56 & 99.58 \\
                        \hline
                    \end{tabular}
                    \label{MlModels}
            \end{center}
        \end{table*}

    Initially, We have employed Random Forest, Decision Tree, K-Nearest Neighbor(KNN), Support Vector Machine(SVM), Multinomial Naive Bayes, Logistic Regression, Stochastic Gradient Descent(SGD). and later we further employed BERT with Stratified K Fold to evaluate our system. BERT is an artificial intelligence (AI) method for comprehending natural language and is also known as Bidirectional Encoder Representations from Transformers. In Transformers, each output element is connected to each input element, and the weightings between them are dynamically determined based on their connection. It only requires the encoder component because its goal is to build a language representation model. The BERT encoder receives a sequence of tokens, which are then translated into vectors and processed by the neural network. The Transformer works by stacking a layer that maps sequences to sequences, resulting in an output that is also a set of vectors with a 1:1 correlation between input and output tokens at the same index. In a number of tasks relevant to general language understanding, such as natural language inference and linguistic acceptability, BERT outperformed the present state-of-the-art. To fine-tune the model, we must first transform the data into the particular format used for pre-training the core BERT models, which includes adding special tokens to label the beginning and end of sentences, as well as segment IDs to differentiate different sentences, and then transform the data into the features used by  BERT. Moreover, K-fold is a cross-validation approach used to measure a machine learning model's skill using unknown data. It is often used to validate a model since it is simple to understand and execute, and the findings are more useful than traditional validation methods.
    
    %Stratified K-Folds To divide data into train and test sets, the cross-validator gives train/test indices. A K-Fold variant that returns stratified folds is this cross-validation object. The folds are created by keeping track of the sample percentages for each class.
    
    %An open source machine learning framework for NLP is called BERT. By establishing context with other text, it is intended to assist computers in deciphering unclear language in text. With the help of question and answer datasets, the BERT framework can be adjusted after being pre-trained on text from Wikipedia. 

        \begin{figure}[!t]
            \centering
            \includegraphics[scale= .5]{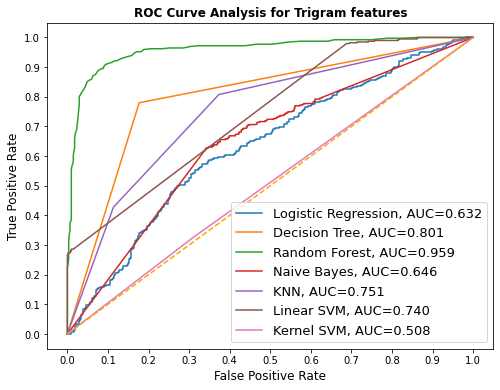}
            \caption{Traditional Machine Learning Algorithms: ROC Curve}
            \label{MLROC1}
        \end{figure}
        
        \begin{figure}[!t]
            \centering
            \includegraphics[scale= .5]{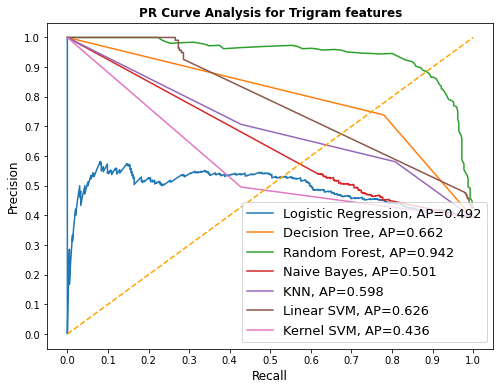}
            \caption{Traditional Machine Learning Algorithms: Precision-Recall Curve}
            \label{MLROC2}
        \end{figure}

        \begin{figure*}[!t]
            \centering
            \includegraphics[scale= .44]{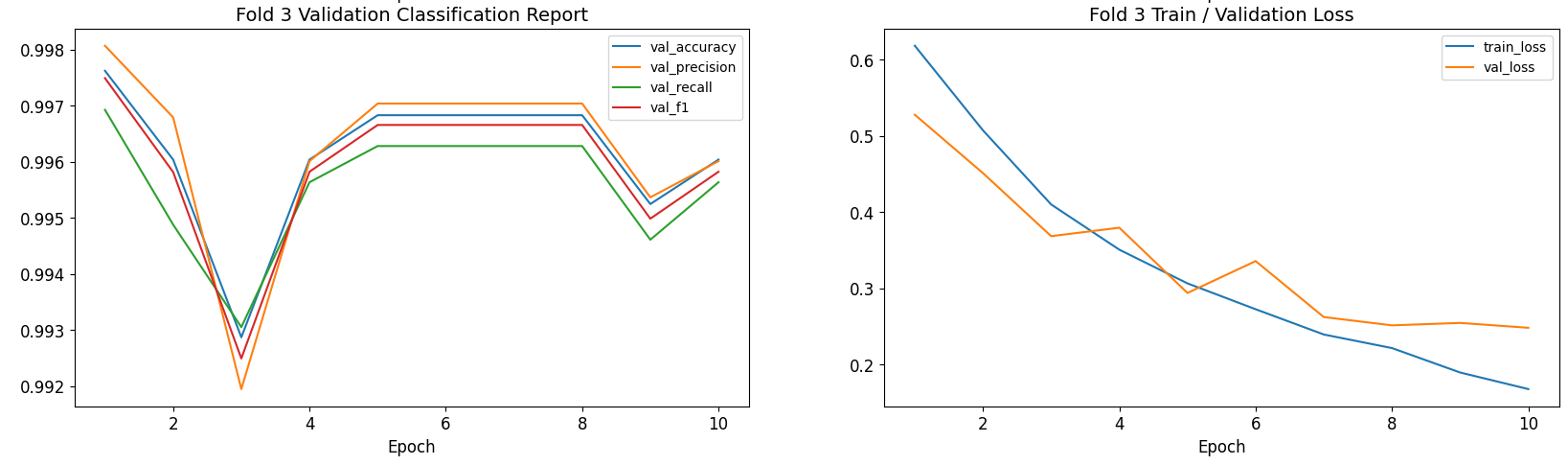}
            \caption{BERT using Stratified K-Fold 4: Left Image depicts the curve of validation Accuracy, Precision, Recall, and F-1 Score. The right image represents the loss curve during the training.}
            \label{proposedmodelFold4}
        \end{figure*}
        
        \begin{figure*}[!t]
            \centering
            \includegraphics[scale= .56]{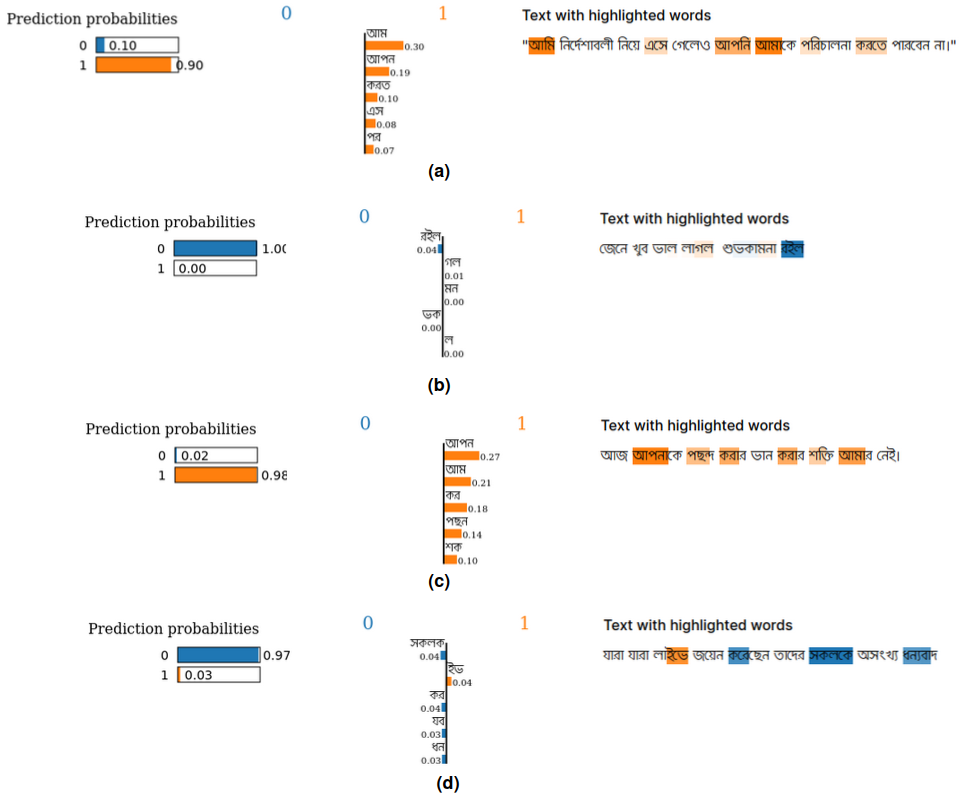}
            \caption{LIME Prediction: Here, Image (a), (b), (c), and (d) are sample inputs. Input (a) and (c) are classified as Sarcasm. (b) and (d) is predicted as non-sarcasm. Orange highlighted regions represent the negative or sarcasm level of a single word whereas Blue highlighted regions are represented as positive or non-sarcasm.}
            \label{fig:x LimeExplaiin}
        \end{figure*}
        
    % \begin{table*}[!t]
    %     \caption{Performance Evaluation: BERT using Stratified K-Fold}
    %         \begin{center}
    %                 \begin{tabular}{|c |c |c |c |c |c |c |c |}
    %                     \hline
    %                     Model  & Precision & Recall &  F-1 Score & Validation & Validation & Validation & Validation \\ [0.5ex]
    %                       &  &  &  & Accuracy & Precision & Recall & F-1 Score\\
    %                     \hline
    %                     Fold 0  & 95.26 & 95.66 & 95.45 & 90.58  & 90.06  & 90.09 & 90.07 \\
    %                     \hline
    %                     Fold 1  & 99.25 & 99.07 & 99.16 & 95.88  & 95.59  & 95.73 & 95.66 \\
    %                     \hline
    %                     Fold 2  & 99.75  & 99.63 & 99.69 & 98.34 & 98.41  & 98.07  & 98.24 \\
    %                     \hline
    %                     Fold 3  & 99.83  & 99.77 & 99.80 & 99.60  & 99.60  & 99.56 & 99.58 \\
    %                     \hline
    %                 \end{tabular}
    %                 \label{BERT}
    %         \end{center}
    %     \end{table*}

\section{Performance Evaluation AND ANALYSIS}
    
 %\subsection{Evaluation Metrics} \label{metrics}

%\subsection{Performance Evaluation} \label{modelPerformance}    
We have evaluated our dataset using our proposed system. We have divided our system training into two parts. In the initial part, we have employed various machine learning algorithms which is discussed in part \ref{TML}, and in the later part we have utilized BERT using Stratified K-Fold which is discussed in part \ref{BERT_Kfold}.
    \subsection{Traditional Machine Learning Algorithms} \label{TML}
         Table \ref{MlModels} depicts our findings employing traditional machine learning algorithms. We have employed  Random Forest, Decision Tree, K-Nearest Neighbor(KNN), Support Vector Machine(SVM), Multinomial Naive Bayes, Logistic Regression, and Stochastic Gradient Descent(SGD). Among these algorithms, Random Forest managed to acquire an accuracy of 91.10\% with a precision, recall, and F-1 score of 89.82\%, 87.09\%, and 88.43\% respectively. Decision Tree however managed to reach 80.61\% with precision, recall, and F-1 score between 73.86\% to 77.97\%. The rest of the algorithms performed poorly with an accuracy ranging from 71\% - 60\%. Figure \ref{MLROC1} represents the corresponding ROC curve for all 7 of the mentioned machine learning framework. Here, we can visualize the superiority of Random forest among the other frameworks. In addition, Figure \ref{MLROC2} depicts the precision-recall curve for the mentioned models. 
        
        %Our dataset were used to test our system. Using our proposed framework, we were able to comprehend and classify a prediction. We began by training multiple models on our dataset. Table II shows the results of this test. For comparison, we used seven machine learning algorithms. Except for Multinomial Naive Bayes, Logistic Regression, and Stochastic Gradient Descent, the majority of the algorithms performed well in this case. Stochastic Gradient Descent had the lowest accuracy on our dataset, while Random Forest had a decent accuracy. The performance of the Decision Tree and Random Forest models was average. Random Forest scored 91 percent, Decision Tree scored 80 percent, and K-Nearest Neighbor and Support-Vector Machines scored 71 percent on our dataset.

    \subsection{BERT} \label{BERT_Kfold}
        Table \ref{MlModels}'s last row represents our findings of our training using BERT with Stratified K-fold. At the initial fold, our model already managed to reach a validation accuracy of 90.58\% with a validation precision, recall, and F-1 score of  90.06\%, 90.09\%, and 90.07\% respectively. This proves the stability of the model's training. In the next fold, it's accuracy improved to 95.88\% which is already superior to traditional machine learning algorithms. In terms of its validation precision, recall, and F-1 score it again scored around 95\% in each section. In Fold 3, it was still able to maintain its increasing accuracy. Finally, in Fold 4 which is presented in the table, we get a validation accuracy of 99.60\% with precision, recall, and F-1 score ranging from 99.56\% to 99.60\%. Figure \ref{proposedmodelFold4} depicts the accuracy, validation, recall, and F1-score curve along with its train loss and validation loss curve.

    \subsection{Local Interpretable Model-Agnostic Explanations}
        We have employed LIME which introduces explainability to our system. Figure \ref{fig:x LimeExplaiin} shows 4 inputs and their prediction along with LIME prediction. Here, Input (a) is confidently identified as sarcasm, and LIME highlights the main elements so that the users may see how LIME arrived at its conclusion. However, Input (b) is predicted as non-sarcasm with 100\% accuracy and the reason behind its prediction is highlighted as well. While Interpreting Input (c), our proposed system furthermore made a confident prediction with  98\% accuracy. It was predicted as sarcasm. Finally, Input (d) was firmly identified as a non-sarcasm, and LIME has emphasized both the positive and negative qualities.

\section{Conclusion}
In this article, we have proposed a BERT-based model for detecting Bangla sarcasm that is capable of achieving 99.60\% whereas utilizing the traditional machine learning models reached only up to 89.93\% accuracy. Another novel contribution of this study is the utilization of a new Bangla sarcasm dataset, BanglaSarc that was collected specifically for the  evaluation of this study's proposed system. Furthermore, we have utilized LIME which introduces explainability to our system. BERT's pre-trained layer along with the Fully connected layer is utilized in this study to grasp deep features and classify sarcasm. Additionally, we have employed various machine learning algorithms and conducted a thorough comparison in order to find the best-fitted model. NLP-based technologies are gaining popularity recently and need cutting-edge accuracy to function properly. However, sarcasm becomes a significant issue since it is often difficult to detect and  thus reduces the overall system's performance. Although, numerous work has been conducted in detecting sarcasm from the English language, very low effort has been put into detecting sarcasm from the Bangla language. Our proposed approach would address this issue and bridge the gap between these two languages in terms of sarcasm identification. In the future, we hope to increase the quantity along with the quality of the dataset. Moreover, we aim to improve memory efficiency along with inference time by proposing a lighter framework.

\vspace{12pt}

\end{document}